\documentclass[journal]{IEEEtran}
\usepackage{color}
\usepackage{algorithm2e}
\usepackage{algorithmic}
\usepackage{pgfplots}
\usepackage{amsfonts}
\usepackage{amssymb}
\usepackage{mathtools}
\usepackage{bbm}
\usepackage{dsfont}
\usepackage{float,comment}

\ifCLASSINFOpdf
\else
\fi
\begin{document}
%
\title{ReLGAN: Generalization of Consistency for GAN with Disjoint Constraints and Relative Learning of Generative Processes for Multiple Transformation Learning}
%
%
%

\author{Chiranjib~Sur \\ 
		Computer \& Information Science \& Engineering Department, University of Florida.\\
		Email: chiranjibsur@gmail.com
}

%
%

\markboth{Journal of XXXX,~Vol.~XX, No.~X, AXX~20XX}%
{Shell \MakeLowercase{\textit{et al.}}: Bare Demo of IEEEtran.cls for IEEE Journals}
%

\maketitle

\begin{abstract}
Image to image transformation has gained popularity from different research communities due to its enormous impact on different applications, including medical. In this work, we have introduced a generalized scheme for consistency for GAN architectures with two new concepts of Transformation Learning (TL) and Relative Learning (ReL) for enhanced learning image transformations. Consistency for GAN architectures suffered from inadequate constraints and failed to learn multiple and multi-modal transformations, which is inevitable for many medical applications. The main drawback is that it focused on creating an intermediate and workable hybrid, which is not permissible for the medical applications which focus on minute details. Another drawback is the weak interrelation between the two learning phases and TL and ReL have introduced improved coordination among them. We have demonstrated the capability of the novel network framework on public datasets. We emphasized that our novel architecture produced an improved neural image transformation version for the image, which is more acceptable to the medical community. Experiments and results demonstrated the effectiveness of our framework with enhancement compared to the previous works. 
\end{abstract}


%
\IEEEpeerreviewmaketitle

\section{Introduction} \label{section:introduction}
\begin{figure*}[!t]
\centering
\includegraphics[width=.8\textwidth]{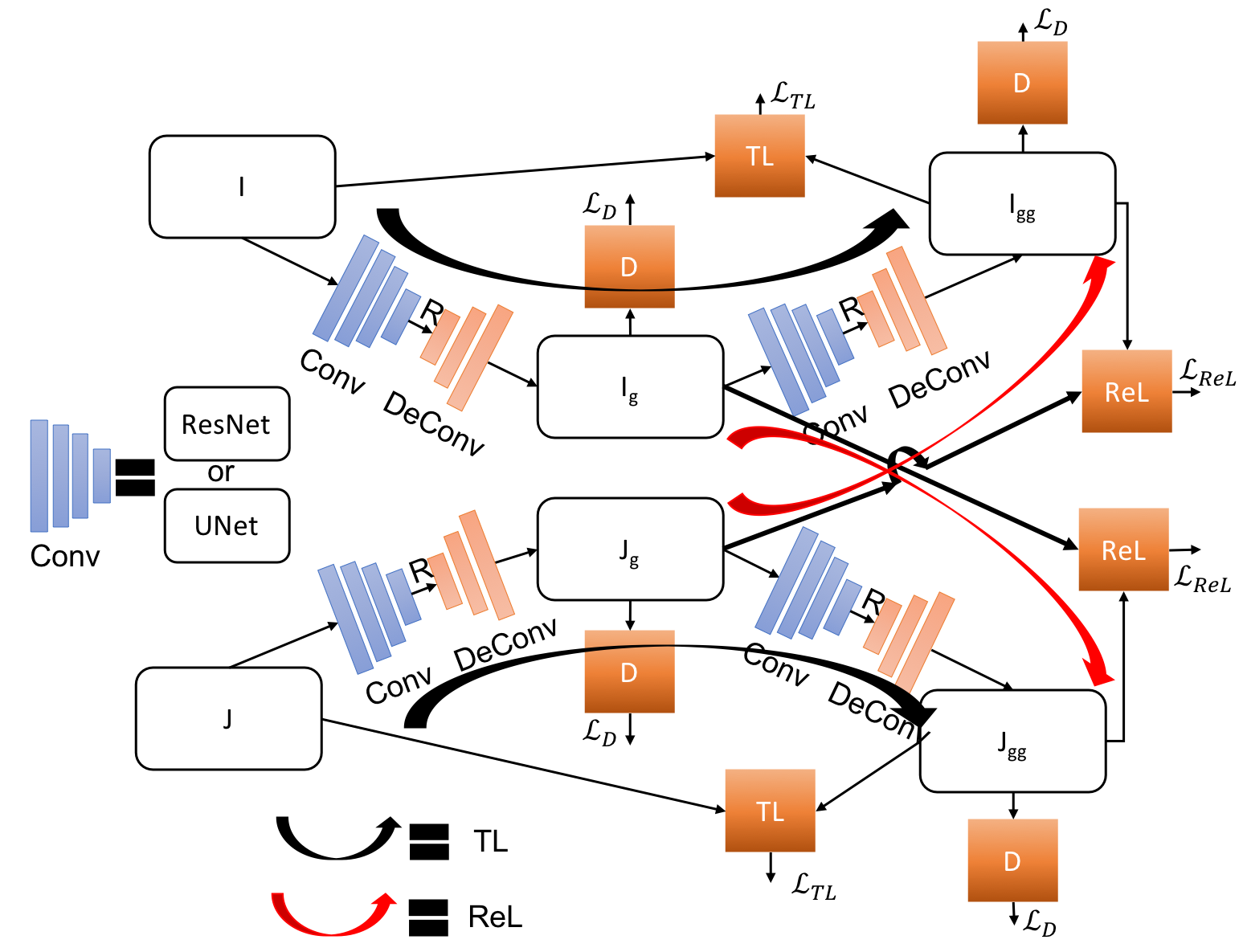}
\caption{A diagrammatic overview of ReLGAN with disjoint constraints are illustrated through channels of (black and red) arrows.} \label{fig:arch}
\end{figure*}

\begin{figure*}[!t]
\centering
\includegraphics[width=.7\textwidth]{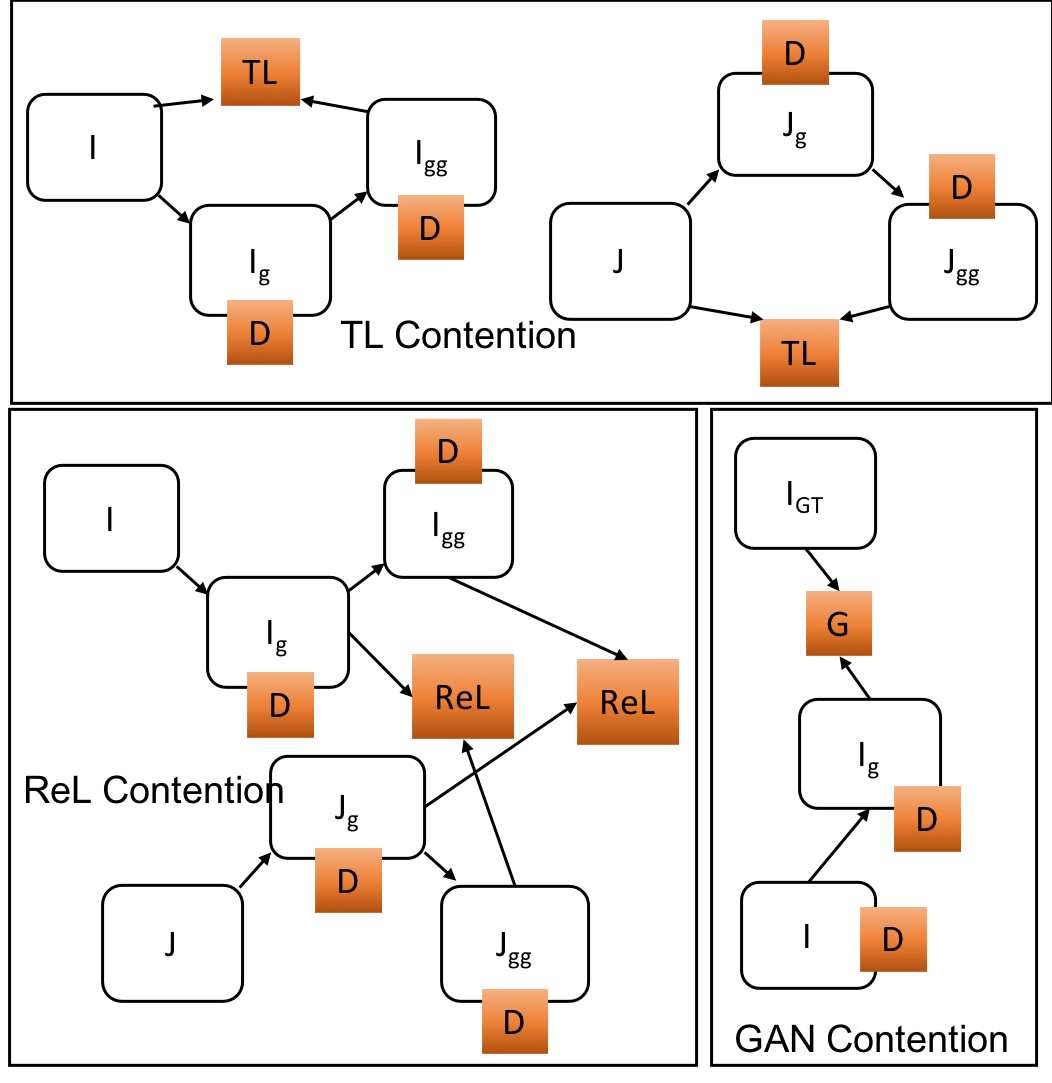}
\caption{An illustration of our architecture with Transformation Learning (TL) Contention and Relative Learning (ReL) Contention for ReLGAN, compared with GAN Contention.} \label{fig1:contention}
\end{figure*}

What Claude Monet imagined was fancier than what we need for enabling the life of a massive number of patients all over the world. Learning translation is challenging. Previous works in image-2-image translations \cite{isola2017image}, \cite{liu2017unsupervised} were focused on mere feature translation. But what if the problem is more complex like detection of organs and painting them with color, instead of using segmentation. Or the texture of the image needs to be reconfigured, keeping intact the feature composition and preventing any foreign object introduction. 
\cite{qu2019enhanced}, \cite{bahrami2016convolutional}, \cite{qu2020synthesized} has provided some examples of such kinds of medical image scenarios, which can make devices more sophisticated and smarter. There are more significant challenges in these kinds of image-2-image translation problems, and we have introduced a new architecture that can handle them. 
Also, multiple translation learning can be challenging and requires extra attention. We have introduced new learning strategies that can enhance performance. 
We have adopted the baseline of CycleGAN \cite{zhu2017unpaired}, \cite{yi2017dualgan} architecture for our experiments.  Our model has the natural flow of learning the translation in both ways, which can be very useful in many medical image translations. 
We can define the problem as $G_{AB}: \textbf{A} \rightarrow \textbf{B} $, $ G_{BA} : \textbf{B} \rightarrow \textbf{A} $ for $\textbf{A} \in S, \textbf{B} \in S$ for CycleGAN \cite{zhu2017unpaired}, \cite{yi2017dualgan}. 
Pairing training set generates additional constraint and can be defined as $G_{AB}: \textbf{A} \rightarrow \textbf{B} $, $ G_{BA} : \textbf{B} \rightarrow \textbf{A} $ for $(\textbf{A},\textbf{B}) \in S$.  
We have generalized the constraints as Transformation Learning (TL) and is defined as $G_{AB}: \textbf{A} \rightarrow \textbf{B}' $, $ G_{BA} : \textbf{B} \rightarrow \textbf{A}' $, $G_{AB}': \textbf{A}' \rightarrow \textbf{B}'' $, $ G_{BA}' : \textbf{B}' \rightarrow \textbf{A}'' $. When the generators are separated individual, it solidifies the learning, though a few of them will be useful.
TL operates on the contention between generation, discrimination, and multiple translations. But after some training sessions, the learning gradient saturates because of the model limitations to differentiate between background texture and the features. Hence, we introduce TL for feature capturing and Relative Learning (ReL) for understanding the difference between background texture. Figure \ref{fig:arch} provided a diagrammatic illustration of the new scheme.

ReL introduces these two new utilities that can enhance the learning process. 
First is a new contention scheme for enhanced learning (without over-fitting) and generalization of representations.
Secondly, after the features are translated, relative learning helps in learning the background. 
For most image-2-image translation, the features are the main translated objects, but ReL involves learning more about the background as well. This is the reason why our novel architecture ReLGAN performs much better than the previous architectures. We have shown through zebra-horse dataset that our architecture is far better and outperformed CycleGAN. 
ReL also involves a smoothening process in comparison to the TL, which introduces sharp gradients between the generated images and the ground-truth images.
In GAN \cite{gupta2019gan} architectures, multiple translations are independent learning strategies that do not generalize well.  But, ReL improves in coordination and cooperation among multiple generators for enhanced generalization of the representations. 
While TL helps in the determination of the critical feature transformation, ReL helps in generalization through comparison of the relative generation. ReL is learning from the relative knowledge of two generators. ReL converges only if there is generalization among the generators. Compared to multiple loss based learning models, which is dependent on TL, ReL is a space that can be defined as a latent space contributed by all different translations and hence generalization of consistencies for any GAN model. 

The rest of the document is arranged with 
ReLGAN architectural details in Section \ref{section:architecture},  details of implementation and methodology, analysis of the experiments and results in Section \ref{section:results}, concluding remarks with future works in Section \ref{section:conclusion}, and two pages of Supplementary Material.

The main contributions of this work are: 1) ReLGAN with Relative Learning strategy for enhanced image-to-image translation 2) The notion of disjoint constraints in Transformation Learning and Relative Learning 3) applied to public dataset, outperformed CycleGANThe main contributions of this work are: 1) ReLGAN with Relative Learning strategy for enhanced image-to-image translation 2) The notion of disjoint constraints in Transformation Learning and Relative Learning 3) applied to public dataset, outperformed CycleGAN.

The rest of the document is arranged with 
the details and scope of the problem in Section \ref{section:problem}, 
the description and statistics of the data in Section \ref{section:data},
Section \ref{section:},
Section \ref{section:},
Section \ref{section:},
Section \ref{section:},
Section \ref{section:},
the intricacies of our methodology in Section \ref{section:methodology}, 
experiments, results and analysis in Section \ref{section:results},
revisit of the existing works in literature in Section \ref{section:literature},
concluding remarks with future prospects in Section \ref{section:discussion},

\section{ReLGAN Architecture} \label{section:architecture}
Relative Learning based Generative Adversarial Network (ReLGAN) works on the principle of contention based learning like any Generative Adversarial Network (GAN), but with enhanced contention strategies. GAN has learning limitations due to the contention exhaustion of the discriminator and the generator. To improve learning, we have introduced ReL based contention for the first time in the literature. Relative because Relative Learning based gradient is much smoother than the absolute difference based gradient. ReL participates as a natural smoothening gradient for the model training.

Considering the generator for the transformation $A \rightarrow B$ as $\text{G} _{_{A \rightarrow B}}(.)$ and for $B \rightarrow A$ as $\text{G} _{_{B \rightarrow A}}(.)$, $\text{E}(.)$ as encoder, $\text{D}_A$, $\text{D}_B$ is the cumulative discriminative losses for both $\textbf{A}$ and $\textbf{B}$ respectively, we define the notions for defining the equations below. Compared to previous works like CycleGAN \cite{zhu2017unpaired}, \cite{yi2017dualgan}, Cycle Consistency Loss is the sole constraint with the equation below.  
\begin{equation}
 L_{C} = \mathop{\mathbb{E}}_{\substack{\textbf{A} \sim \text{Pr}(\textbf{A}) \\ \textbf{B} \sim \text{Pr}(\textbf{B})}} [ \parallel \textbf{A} - \text{G} _{_{B \rightarrow A}}( \text{G} _{_{A \rightarrow B}} (\textbf{A}) ) \parallel] 
\end{equation}
Augmented CycleGAN \cite{almahairi2018augmented} was introduced with encoder based random variable tensor. Here, encoder based distribution learning ($z \approx \text{E}(.)$) in Consistency Loss is constrained as the following.
\begin{equation}
 L_{AC} = \mathop{\mathbb{E}}_{\substack{\textbf{A} \sim \text{Pr}(\textbf{A}) \\ \textbf{B} \sim \text{Pr}(\textbf{B})}} [ \parallel \textbf{A} - \text{G} _{_{B \rightarrow A}}( \text{E}(\textbf{B}), \text{G} _{_{A \rightarrow B}} (\text{E}(\textbf{A}), \textbf{A}) ) \parallel] 
\end{equation}

But ReLGAN is about more generalized constraints and for multiple levels or hierarchical and disjoint losses of the cycles. 
We define Transformer Learning (TL) (same as Cycle Consistency Loss but with independent generators) for learning of the transformation and Relative Learning (ReL) for improving the quality of the transformation. Figure \ref{fig:arch}-\ref{fig1:contention} explains the architecture. Apart from learning the difference with the ground-truth, the key is bridging the gap between different stages as constraints and thus can produce much better feature capturing capability. These two kinds of constraints are very important for medical image transformation as it helps capture minute details and also important aspects that cannot be considered as fancy as an intermediate hybrid. We define the equations for ReLGAN as the following,
\begin{equation} \label{eq:ReL}
\begin{split}
 & L_{GC}  = \mathop{\mathbb{E}}_{\substack{\textbf{A} \sim \text{Pr}(\textbf{A}) \\ \textbf{B} \sim \text{Pr}(\textbf{B})}} [ \text{ } \text{ }   \parallel \textbf{A} - \text{G} _{_{B \rightarrow A}}( \text{G} _{_{A \rightarrow B}} (\textbf{A}) ) \parallel  \\
 & + \parallel \textbf{B} - \text{G} _{_{A \rightarrow B}}(\textbf{A}) \parallel  + \parallel \text{G} _{_{A \rightarrow B}}(\textbf{A}) - \text{G} _{_{A \rightarrow B}}( \text{G} _{_{B \rightarrow A}}(\textbf{B}))  \parallel \text{ } ] 
 \end{split}
\end{equation}
Here, $\parallel \textbf{A} - \text{G} _{_{B \rightarrow A}}( \text{G} _{_{A \rightarrow B}} (\textbf{A}) ) \parallel$ is the Transformation Loss for Learning, $\text{ReL}_1 = \parallel \textbf{B} - \text{G} _{_{A \rightarrow B}}(\textbf{A}) \parallel $ is Relative Loss for Learning  and $\text{ReL}_2 = \parallel \text{G} _{_{A \rightarrow B}}(\textbf{A}) - \text{G} _{_{A \rightarrow B}}( \text{G} _{_{B \rightarrow A}}(\textbf{B}))  \parallel$ is another Relative Loss for Learning. Initially, $\text{ReL}_1$ is dominant until stagnant, when $\text{ReL}_2$ is introduced for further contention.  Later $\parallel \text{G} _{_{A \rightarrow B}}(\textbf{A}) - \text{G} _{_{A \rightarrow B}}( \text{G} _{_{B \rightarrow A}}(\textbf{B}))  \parallel$ is effective to prevent over-fitting and continues the learning. At one point in GAN \cite{gupta2019gan}, the discriminator stops the learning because of its enhanced ability to discriminate the generated samples from the ground-truth. Additional constraints in ReL are $\parallel \textbf{B} - \text{G} _{_{A \rightarrow B}}(\textbf{A}) \parallel$ and $\parallel \text{G} _{_{A \rightarrow B}}(\textbf{A}) - \text{G} _{_{A \rightarrow B}}( \text{G} _{_{B \rightarrow A}}(\textbf{B}))  \parallel$ as demonstrated in Figure \ref{fig1:contention}. 
Some researchers argue that $\parallel \textbf{B} - \text{G} _{_{A \rightarrow B}}(\textbf{A}) \parallel$ is the same old MSE loss, but we argue that this loss $\text{ReL}_1$ helps in learning the relative loss, while $\parallel \text{G} _{_{A \rightarrow B}}(\textbf{A}) - \text{G} _{_{A \rightarrow B}}( \text{G} _{_{B \rightarrow A}}(\textbf{B}))  \parallel$ drags the generated version of $\textbf{B}$ to a better sub-space, until generalization and identification of the necessaries and the minute details. Here, $G_{XY} \equiv G_{X \rightarrow Y}$.
Overall ReLGAN loss function is denoted as, 
\begin{equation}
\begin{split}
 & \mathcal{L}(D_B, D_A, T_L, R_{eL}) = \mathop{\mathbb{E}}_{\textbf{B} \sim \text{Pr}(\textbf{B})} [ \log D_B(\textbf{B}) ]  
  \\
 &+ \mathop{\mathbb{E}}_{\textbf{A} \sim \text{Pr}(\textbf{A})} [ \log (1 - D_B(G_{AB}(\textbf{A})))]  + 
 \mathop{\mathbb{E}}_{\textbf{A} \sim \text{Pr}(\textbf{A})} [ \log D_A(\textbf{A}) ]  
  \\
 &+ \mathop{\mathbb{E}}_{\textbf{B} \sim \text{Pr}(\textbf{B})} [ \log (1 - D_A(G_{BA}(\textbf{B})))] \\
 & + 
 \mathop{\mathbb{E}}_{\textbf{A} \sim \text{Pr}(\textbf{A})} [ ( \parallel \textbf{A} - G_{BA}'(G_{AB}(\textbf{A})) \parallel )] 
  \\
 &+
 \mathop{\mathbb{E}}_{\textbf{B} \sim \text{Pr}(\textbf{B})} [ ( \parallel \textbf{B} - G_{AB}'(G_{BA}(\textbf{B})) \parallel )] \\
 & + 
 \mathop{\mathbb{E}}_{\substack{\textbf{A} \sim \text{Pr}(\textbf{A}) \\ \textbf{B} \sim \text{Pr}(\textbf{B})}} [ ( \parallel G_{BA}'(G_{AB}(\textbf{A}))  - G_{}(\textbf{B}) \parallel )] 
 \\
 &+ 
 \mathop{\mathbb{E}}_{\substack{\textbf{A} \sim \text{Pr}(\textbf{A}) \\ \textbf{B} \sim \text{Pr}(\textbf{B})}} [ ( \parallel G_{AB}'(G_{BA}(\textbf{B}))  - G_{AB}(\textbf{A}) \parallel )]
\end{split}
\end{equation}

\subsection{Translation Learning} 
Transformation Loss based learning works much better than mean square based generative loss (like in Auto-Encoder, GAN) because of its steep gradient and it cannot learn the latent transformed space for images. We are more interested in the CycleGAN based transformations of both directions as it captures hidden transformation truths that are beyond simple translation problems like denoising \cite{tripathi2018correction}, segmentation \cite{gupta2019gan} and super-resolution \cite{wang2018esrgan}. 
Translation Learning can be defined mathematically as the following equation,
\begin{equation}
 G_{AB}: \textbf{A} \rightarrow \textbf{B}' \text{ , }  G_{BA} : \textbf{B} \rightarrow \textbf{A}' \text{ , }  G_{AB}': \textbf{A}' \rightarrow \textbf{B}'' \text{ , } G_{BA}' : \textbf{B}' \rightarrow \textbf{A}''
\end{equation}
This equation for TL is a generalized Cycle-Consistency Loss and imposes disjoint constraints for learning representations. 

\subsection{Disjoint Constraints} 
Differentiating $G_{AB}$ with $G_{AB}'$ and $G_{BA}$ with $G_{BA}'$ have  provided significant scope for improvement and were not present in CycleGAN. The main reason is the disjoint constraints like $ G_{AB}'(G_{BA}(\textbf{B})) \rightarrow \textbf{B} $ and $ G_{BA}'(G_{AB}(\textbf{A})) \rightarrow \textbf{A} $ instead of $ G_{AB}(G_{BA}(\textbf{B})) \rightarrow \textbf{B} $ and $ G_{BA}(G_{AB}(\textbf{A})) \rightarrow \textbf{A} $ in CycleGAN. 
ReL is also part of the notion of disjoint constraints as more constraints are imposed for generalization, as shown Equation \ref{eq:ReL}, instead of relying on TL. 
ReL helps in information exchange between the cycle and also in making sure that the learning is reciprocating among different parts of the network. TL and ReL are disjoint constraints because they manage the learning phases of two different criteria. In Transformation Learning, data $A$ is considered, while in Relative Learning, data $B$ is considered and vice-versa for the other cycle.

\subsection{Relative Learning} 
Relative Learning Generative Adversarial Network (ReLGAN) initially uses Transformation Learning (TL) and later fine-tunes with Relative Learning (ReL), which has a relative finer gradient due to the relative knowledge comparison between two generators. For an enhanced transformation, it is very important to impose constraints. Apart from learning the TL loss for transformation, we introduced the novel idea of ReL to impose more constraints on the model, make the different transformation interdependent, and extract more generalization information. 
Relative Learning can be defined mathematically as the following equation,
\begin{multline}
 \mathcal{L}_{AB}^{ReL}: G_{AB}' \text{ , } G_{AB} \equiv \text{Pr}(B') \approx \text{Pr}(B'') \text{ and } \\
 \mathcal{L}_{BA}^{ReL}: G_{BA}' \text{ , } G_{BA} \equiv \text{Pr}(A') \approx \text{Pr}(A'')
\end{multline}
ReL will be enhance different translation learning as medical applications are getting more delicate and models need to learn very minute details. Medical applications are sensitive to generative processes and require more control over the features. ReL will provide such improved controls.


\section{Results \& Discussion} \label{section:results}
Image translation problems like denoising \cite{tripathi2018correction}, segmentation \cite{gupta2019gan} and super-resolution \cite{wang2018esrgan} are spatially proportional/similar translation and much easier to solve like pseudo-conditional probability $\text{Pr}(\textbf{A}+\delta(\textbf{A}) \mid \textbf{A})$. We are trying to solve problems like removing haziness \cite{qu2019enhanced} in images, or perform some non-conventional transformations like 3T to 7T MRI \cite{qu2020synthesized} or locating a specific organ and provide a clearer and detailed view or true conditional probability $\text{Pr}(\textbf{B}\mid \textbf{A})$.

\subsection{Dataset Description} 
To demonstrate the effectiveness of our strategy we have chosen zebra-horse translation dataset and performed a comparison with CycleGAN \cite{zhu2017unpaired}, \cite{yi2017dualgan}. 
Learning the latent distribution and intermediates, unsupervised, can be challenging. But, to learn properties like locating the horse and not disturb the non-horse brown structures, the task is relatively tricky. Our proposed joint learning method contributed to this conceptual framework and produced promising results. 
While two transformations can be easier to train but are lengthy, and the inference is fast. Experiments were done in Tensorflow in K80 Tesla GPU with 32 GB RAM with image dimension $256 \times 256$. We used the training-testing split of the dataset as it is.

\subsection{Experiments \& Results} 
The model is trained end-to-end for different horse-zebra combination subset for several epochs. Later, the test set is generated, and some instances are provided in Figure \ref{fig1:results_ReLGAN} and more in Supplementary Material. In the absence of any particular ground-truth, other numerical metrics like SSIM cannot be reported, but a visual inspection will surely conclude that ReLGAN outperformed CycleGAN. ReLGAN also performed much better than CycleGAN for many internal datasets for an organization. 

There are severe drawbacks in CycleGAN, mainly when the generators try to learn both the transformation, where one is inverse of the other. However, these latent representations are more driven towards the inverse generation consistency,  leading to hybridization, and not complete learning of the usable intermediate spaces. CycleGAN's inverse constraint strategy fails for cross-domain translations. \cite{almahairi2018augmented} proposed a transformed random variable for style encoding, which is difficult to define for medical images. In \cite{almahairi2018augmented} situations, learning both way translations are beneficial as paired data converges to a definite random variable based latent space. In fact, many aesthetic applications can tolerate distortions or artifacts. However, in many applications related to health-care and sensitive technologies, the translations and its parity with the ground-truths are more important. 
ReLGAN is driven to incorporate different aspects of the transformation through learning the transformations and yet preventing the dual-direction transformation from getting rid of the chances that can move in either way, which is not desirable.

\begin{figure*}[!t]
\centering
\includegraphics[width=.75\textwidth]{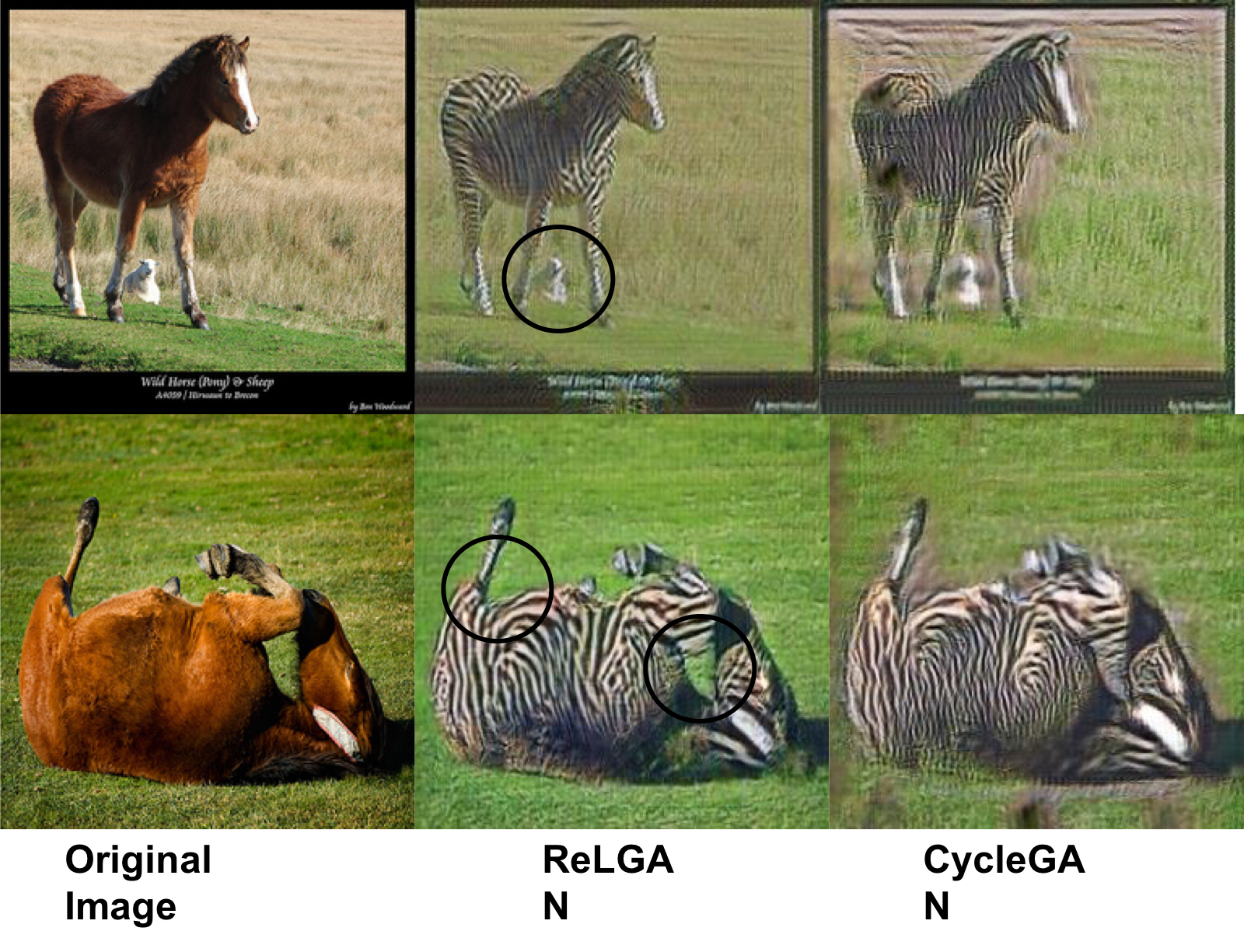}
\caption{Generated Instances with Zebra, For ReLGAN. Supplementary. } \label{fig1:results_ReLGAN}
\end{figure*}

\subsection{Implications of TL and ReL}
Learning joint transformations can be challenging and requires extra sets of information. For cross-domain translation problems, it is very difficult to incorporate all possibilities and complications of a problem. In such a scenario, ReL provides an additional edge over previous works. While TL is an unsupervised fashion of learning and yet to gain full effectiveness in applications related to medical devices, ReL is a smoothening gradient. ReL introduces a  supervised way of learning with comparative knowledge like between two trained generators.  This kind of joint learning of image-to-image transformations for better contrast and feature prominence can help in many medical applications, and in this work, we have provided a model that can help to learn multiples effectively and is part of the strategy to use it extensively in many applications, where it is needed.
Capturing unique characteristics of image collections and configuring these characteristics to be translated into another is the main objective of ReLGAN. In the absence of any paired training samples, this is even more challenging. Some of the highlights of our results are marked with black color ovals in Figure \ref{fig1:results_ReLGAN} and Supplementary materials. It is to be noted that the generated images have horse(s) pointed out, transformed to zebra, and the non-horse parts are left out. CycleGAN distorted some non-horse parts, mainly when they are brown in color. CycleGAN has limitations generating original looking images, which can be a serious issue, mainly in the medical image domain.

\section{Conclusion} \label{section:conclusion}
In this work, we introduced ReLGAN, a novel architecture for image-to-image transformation, mainly for cross-domain and non-conventional translations. ReLGAN is a contention based strategy to learn delicate characteristics of images. With the horse-zebra dataset, the experiments demonstrated that ReLGAN outperformed CycleGAN. ReLGAN is more generalized and gathers minute details that previous models could not. ReLGAN is based on a series of constraints defined as TL and ReL, and these kinds of disjoint constraints are the first of its kind in the literature.

\newpage

\begin{figure*}
\centering
\includegraphics[width=.7\textwidth]{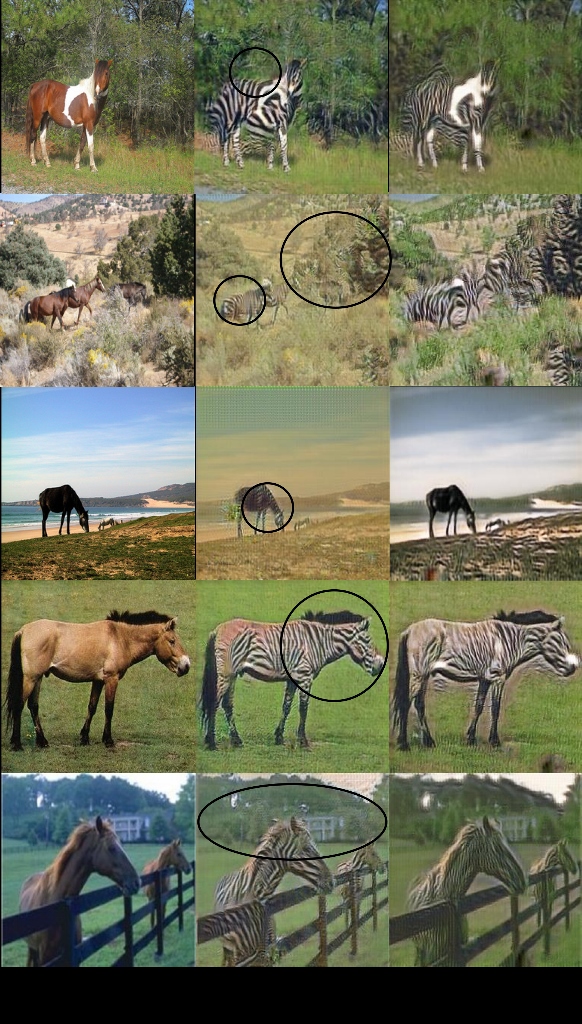}
\label{fig1}
\end{figure*}

\begin{figure*}
\centering
\includegraphics[width=.7\textwidth]{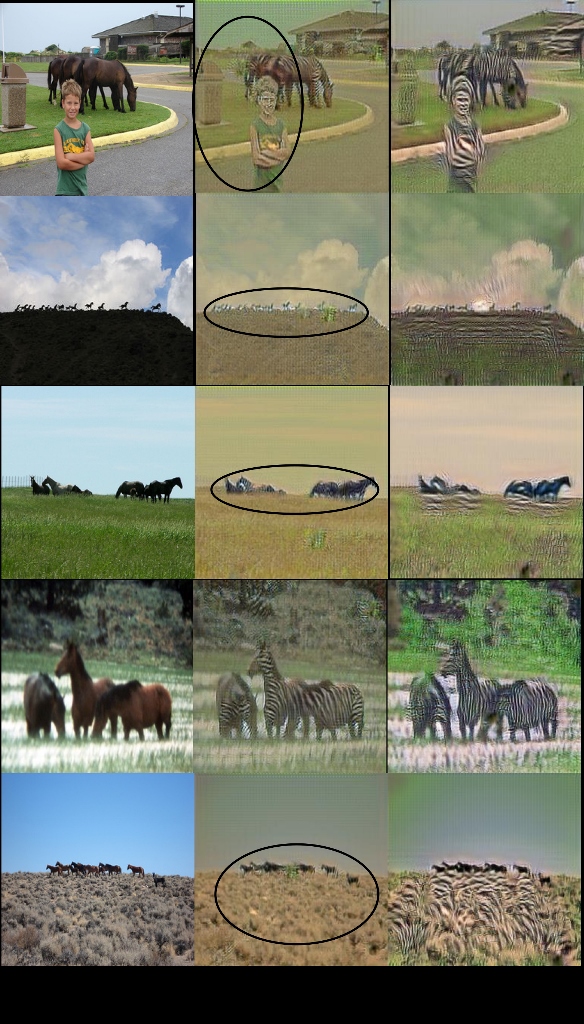}
\label{fig2}
\end{figure*}

\ifCLASSOPTIONcaptionsoff
  \newpage
\fi

%









\begin{thebibliography}{1}



	\bibitem{almahairi2018augmented}
Almahairi, Amjad, et al. "Augmented cyclegan: Learning many-to-many mappings from unpaired data." arXiv preprint arXiv:1802.10151 (2018).
\bibitem{zhu2017unpaired} 
Zhu, Jun-Yan, et al. "Unpaired image-to-image translation using cycle-consistent adversarial networks." Proceedings of the IEEE international conference on computer vision. 2017.
\bibitem{isola2017image} 
Isola, Phillip, et al. "Image-to-image translation with conditional adversarial networks." Proceedings of the IEEE conference on computer vision and pattern recognition. 2017.
\bibitem{liu2017unsupervised}
Liu, Ming-Yu, Thomas Breuel, and Jan Kautz. "Unsupervised image-to-image translation networks." Advances in neural information processing systems. 2017.
  
\bibitem{qu2019enhanced}
Qu, Yanyun, et al. "Enhanced pix2pix dehazing network." Proceedings of the IEEE Conference on Computer Vision and Pattern Recognition. 2019.

\bibitem{bahrami2016convolutional}
Bahrami, Khosro, et al. "Convolutional neural network for reconstruction of 7T-like images from 3T MRI using appearance and anatomical features." Deep Learning and Data Labeling for Medical Applications. Springer, Cham, 2016. 39-47.

\bibitem{qu2020synthesized}
Qu, Liangqiong, et al. "Synthesized 7T MRI from 3T MRI via Deep Learning in Spatial and Wavelet Domains." Medical Image Analysis (2020): 101663.

\bibitem{yi2017dualgan}
Yi, Zili, Hao Zhang, Ping Tan, and Minglun Gong. "Dualgan: Unsupervised dual learning for image-to-image translation." In Proceedings of the IEEE international conference on computer vision, pp. 2849-2857. 2017.

\bibitem{choi2018stargan}
Choi, Yunjey, Minje Choi, Munyoung Kim, Jung-Woo Ha, Sunghun Kim, and Jaegul Choo. "Stargan: Unified generative adversarial networks for multi-domain image-to-image translation." In Proceedings of the IEEE conference on computer vision and pattern recognition, pp. 8789-8797. 2018.

\bibitem{gupta2019gan}
Gupta, Laxmi, Barbara M. Klinkhammer, Peter Boor, Dorit Merhof, and Michael Gadermayr. "GAN-Based Image Enrichment in Digital Pathology Boosts Segmentation Accuracy." In International Conference on Medical Image Computing and Computer-Assisted Intervention, pp. 631-639. Springer, Cham, 2019.
\bibitem{wang2018esrgan}
Wang, Xintao, Ke Yu, Shixiang Wu, Jinjin Gu, Yihao Liu, Chao Dong, Yu Qiao, and Chen Change Loy. "Esrgan: Enhanced super-resolution generative adversarial networks." In Proceedings of the European Conference on Computer Vision (ECCV), pp. 0-0. 2018.

\bibitem{tripathi2018correction}
Tripathi, S., Lipton, Z. C., \& Nguyen, T. Q. (2018). Correction by projection: Denoising images with generative adversarial networks. arXiv preprint arXiv:1803.04477.

\end{thebibliography}
\end{document}